\documentclass[letterpaper, 10 pt, conference]{ieeeconf}  

\IEEEoverridecommandlockouts                              

\usepackage{amsmath} 
\usepackage{amssymb}  
\usepackage{lipsum}
\usepackage{color}
\usepackage[normalem]{ulem}

\usepackage{graphicx}
\usepackage{bm}
\usepackage{microtype}
\usepackage{threeparttable}
\usepackage{booktabs}
\usepackage{multirow}

\usepackage[none]{hyphenat}
\title{\LARGE \bf
Learning-Based Dynamics Modeling and Robust Control for Tendon-Driven Continuum Robots
}

\author{Ziqing Zou$^{*}$, Ke Qiu$^{*}$, Fei Wang, Haojian Lu, Rong Xiong and Yue Wang
\thanks{$^{*}$These authors contributed equally to this work.}%
\thanks{
The authors are with Department of Control Science and Engineering, Zhejiang University, Hangzhou, China.
}}

\begin{document}

\maketitle
\thispagestyle{empty}
\pagestyle{empty}

\begin{abstract}
Tendon-Driven Continuum Robots (TDCRs) pose significant modeling and control challenges due to complex nonlinearities, such as frictional hysteresis and transmission compliance. 
This paper proposes a differentiable learning framework that integrates high-fidelity dynamics modeling with robust neural control. 
We develop a GRU-based dynamics model featuring bidirectional multi-channel connectivity and residual prediction to effectively suppress compounding errors during long-horizon auto-regressive prediction. 
By treating this model as a gradient bridge, an end-to-end neural control policy is optimized through backpropagation, allowing it to implicitly internalize compensation for intricate nonlinearities. 
Experimental validation on a physical three-section TDCR demonstrates that our framework achieves accurate tracking and superior robustness against unseen payloads, outperforming Jacobian-based methods by eliminating self-excited oscillations. 
For implementation details and source code, please refer to https://github.com/ZiqingZou/ContinuumControl.
\end{abstract}

\section{Introduction}
Continuum robots have emerged as a promising solution for navigating unstructured and constrained environments owing to the inherent compliance~\cite{burgner2015survey}, which enables their application in interventional medicine and industrial inspection~\cite{zhang2024copilot}. 
Meanwhile, tendon-driven mechanisms are increasingly integrated into dexterous robotic hands and specialized manipulators~\cite{wen2017tdhand}. 
With bio-inspired architectures, continuum systems overcome the dexterity limitations of traditional rigid-link robots, facilitating delicate and adaptable manipulation in unstructured settings.

However, precise modeling and robust control for Tendon-Driven Continuum Robots (TDCRs) remain a nontrivial task, primarily due to the intricate nonlinearities in physical deformation and tendon actuation. 
On one hand, due to the inherent transmission compliance of elastic tendons and backbone~\cite{amanov2021tendon}, actuator motions are not accurately reflected in the effective tendon length, resulting in significant response latency. 
On the other hand, transmission friction introduces hysteresis and dead-zone nonlinearities~\cite{zhang2023hysteresis}, making feedback control prone to limit-cycle oscillations or divergence, particularly for large inertia or in high-speed tracking. 
Furthermore, the mechanical properties of TDCRs exhibit both spatial and temporal variations. The friction profile often varies with the instantaneous robot configuration and bending history. Additionally, factors such as material creep, viscoelastic deformation, and stress relaxation lead to a gradual shift in both kinematic and dynamic characteristics over time~\cite{jennifer2015elastomers}. 
These uncertainties severely undermine the modeling accuracy and control robustness of TDCRs.

For the modeling, widely adopted kinematic approaches like Piecewise Constant Curvature (PCC)~\cite{webster2010design} typically treat the robot as a quasi-static entity, failing to capture intricate nonlinearities. Conversely, high-fidelity dynamics models such as the Cosserat rod model~\cite{hu2025contact} rely on precise material parameter identification and often face convergence issues in solving nonlinear boundary value problems (BVPs). For the control, Jacobian-based methods provide a linear mapping from task space to actuation space but neglect high-order dynamic characteristics. Furthermore, Model Predictive Control (MPC) is constrained by heavy computational overhead, making it difficult to achieve real-time high-precision performance.

Learning-based approaches have become increasingly viable for the modeling and real-time control of complex nonlinear systems with the advancements of deep learning and accelerated hardware~\cite{liu2025data}. Recurrent Neural Networks (RNNs), with their inherent temporal processing capabilities, are particularly suited for modeling TDCRs and capturing history-dependent nonlinearities and dynamics shifts. However, NN-based auto-regressive prediction is typically plagued by compounding errors. Moreover, the learning of robust control policies via neural networks presents further hurdles. Supervised learning and imitation learning suffer from poor generalization to Out-Of-Distribution (OOD) scenarios, where even minor domain shifts can lead to catastrophic failures~\cite{ross2011dagger}. Conversely, Reinforcement Learning (RL) is hindered by sample inefficiency~\cite{schulman2017ppo, haarnoja2018sac}, where stochastic exploration in the early stages often inflicts irreversible damage on physical TDCR hardware.

In this paper, we propose a differentiable learning framework for the precise modeling and robust control of TDCRs. 
We employ a Gated Recurrent Unit (GRU) as the backbone of our architecture to leverage its efficient gating mechanism in characterizing non-Markovian behaviors. 
To solve the critical challenge of error accumulation in multi-step prediction, we incorporate bidirectional multi-channel connectivity and residual prediction, ensuring stable gradient flow over extended temporal horizons.
Utilizing this differentiable dynamics model as a surrogate environment, we develop an end-to-end neural control policy.
Unlike traditional optimization-based controllers, our policy is optimized offline by backpropagating task-level tracking errors through the dynamics computational graph. 
This allows the policy to implicitly compensate for transmission compliance and friction, achieving robust performance.

Our contributions are summarized as follows:
\begin{itemize}
\item \textbf{High-Fidelity Dynamics Modeling:} We develop an RNN-based architecture featuring bidirectional connectivity and residual prediction to capture TDCR nonlinear dynamics. This design effectively suppresses compounding errors, enabling stable and accurate long-horizon prediction.
\item \textbf{Differentiable Policy Optimization:} We propose a learning pipeline that optimizes the control policy by backpropagating tracking errors through the differentiable dynamics model. This approach internalizes complex nonlinear compensation laws, ensuring efficient real-time execution.
\item \textbf{Experimental Hardware Validation:} We validate our framework on a physical 3-section TDCR, demonstrating superior tracking precision and robustness under external payloads. The results further confirm that our controller effectively suppresses self-excited oscillations common in traditional feedback schemes.
\end{itemize}

\section{Related Work}

\subsection{Continuum Robots Modeling}
Modeling of TDCRs typically involves a trade-off between efficiency and fidelity. 
The PCC model is a widely adopted kinematic model for its simplicity and practicality, offering a rapid mapping from tendon lengths to configuration space~\cite{webster2010design}. However, this circular arc assumption often fails under complex loading or significant deformations~\cite{burgner2015survey}. Although variable curvature models improve shape estimation through modal functions~\cite{renda2020geometric,boyer2020dynamics}, they introduce higher complexity while still neglecting critical nonlinear dynamics, such as transmission compliance and routing friction.
Conversely, the Cosserat rod model provides a geometrically exact description of large elastic deformations and also captures dynamic effects, making it better suited for high-fidelity modeling~\cite{hu2025contact}. Although it can formulate physically rigorous statics via ODEs and dynamics via PDEs~\cite{rucker2011statics, till2019real}, its performance relies on precise identification of material parameters. Furthermore, the high computational cost and convergence issues of solving nonlinear BVPs hinder its application in real-time tasks.

Data-driven approaches are adopted for both kinematics~\cite{grassmann2018learning} and dynamics modeling~\cite{thuruthel2017learning}. 
For articulated continuum robots, RNN-based dynamics learning has been implemented~\cite{hendrik2024rnn}, yet the limited training data necessitates a long warm-up period for hidden state convergence. 
To improve kinematics and dynamics generalization, Physics-Informed Neural Networks(PINNs) leverage first-principle physics~\cite{tim2026pinn}, but they require rigorous calibration and accurate physical parameters.
Meta-learning architectures address dynamic shifts in TDCRs~\cite{tang2026science}, while requiring computationally intensive online parameter updates. 
Our method exhibits dynamic adaptability and maintains accurate multi-step prediction without these limitations.

\subsection{Continuum Robots Control}
Controlling TDCRs remains challenging due to their inherent nonlinearities. 
While Jacobian-based methods provide a linear mapping from task space to actuation space by solving inverse kinematics~\cite{wang2022control, lynch2017modern}, they are often hampered by estimation drifts and singularities. 
Even with analytical Jacobians provided by the PCC model~\cite{qiu2025actuator, gonthina2020mechanics} or numerical Jacobians derived from the Cosserat rod model~\cite{rucker2011computing, campisano2021closed}, these local techniques typically treat the system as quasi-static, thereby neglecting high-order nonlinear dynamics. 
Dynamic controllers have been implemented to mitigate these effects~\cite{della2020model, kazemipour2022adaptive}. 
To consider the future behavior of the robot, MPC has been applied to concentric tube robots~\cite{khadem2020autonomous} and pneumatically actuated continuum robots~\cite{spinelli2022unified}, incorporating multi-step dynamics and system constraints. However, its performance is highly dependent on model fidelity and sensitive to optimization hyperparameters~\cite{hendrik2024rnn}. Moreover, the heavy computational overhead associated with long-horizon optimization makes high-precision and real-time MPC difficult to achieve in practice.

Learning-based approaches offer a promising alternative for the control of continuum robots by bypassing explicit analytical modeling~\cite{satheeshbabu2019open}. Deep learning has been employed to learn inverse kinematics, enabling both tip pose control~\cite{george2017learning, yu2026zhongshan} and full-body shape control~\cite{almanzor2023static}. For dynamic scenarios, data-driven models have been integrated with optimization-based frameworks to achieve open-loop control~\cite{thuruthel2017learning}. Furthermore, MPC has also been implemented using RNNs and PINNs~\cite{hendrik2024rnn, tim2026pinn}, but non-convex optimization complexity limits the look-ahead horizon to fewer than five steps, thus hindering performance in high-speed dynamics tasks. 
Although these methods utilize neural networks as forward models within an optimization loop, there remains a lack of an end-to-end neural control policy that enables both fast inference and robust performance in high-dynamic maneuvers.

\subsection{Differentiable Policy Learning}
Differentiable Policy Learning (DPL) reformulates robot dynamics and control as an end-to-end differentiable computational graph. Unlike "black-box" reinforcement learning methods such as PPO~\cite{schulman2017ppo} and SAC~\cite{haarnoja2018sac}, DPL leverages analytical or numerical gradients to enable direct optimization from task-level loss to policy parameters.
In aerial robotics, agile flight has been optimized through differentiable physics incorporating Newtonian dynamics~\cite{zhang2025flight}. 
Similar principles have been applied to heavy machinery, where high-precision tracking is achieved via differentiable closed-loop dynamics~\cite{zou2025excavator}, with further enhancements through online model-policy co-optimization~\cite{nan2025excavator}. 
In the domain of dexterous manipulation, DPL-based residual policies have been trained to compensate for unmodeled dynamics and bridge the sim-to-real gap~\cite{liu2026dexndm}. 
Our approach extends DPL to the domain of continuum robots, utilizing a differentiable training pipeline to learn a neural control policy that implicitly compensates for nonlinear dynamics and enables precise real-time control.

\section{Dynamics Model}
This section analyzes the intricate transmission characteristics of TDCRs and formulates a learning framework.

\begin{figure}[t]
    \vspace{5pt}
    \centering
    \includegraphics[width=3.3in]{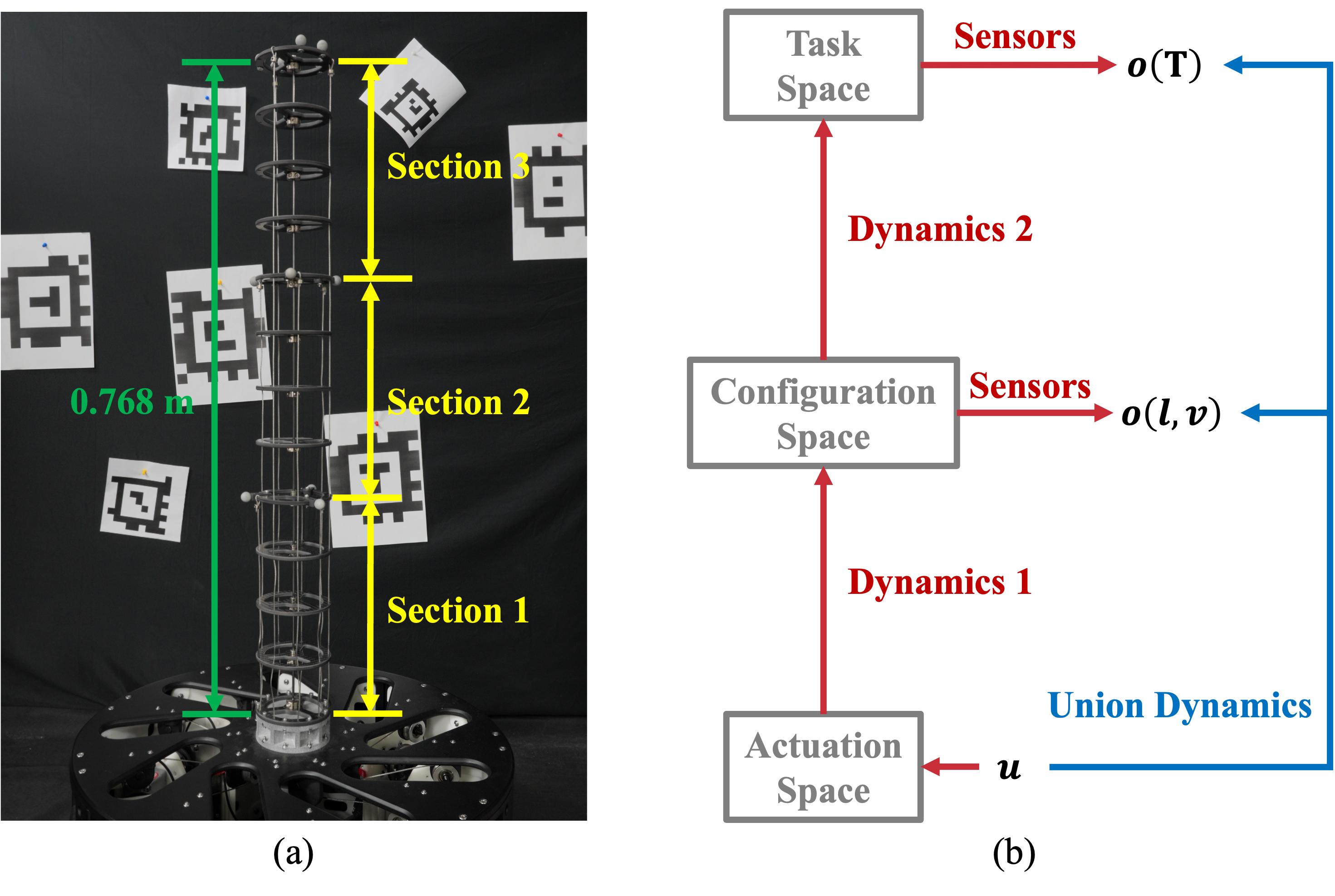}
    \vspace{-10pt}
    \caption{(a) The 0.768m tall three-section TDCR platform used in this study. (b) The system dynamics from control inputs to sensory outputs.}
    \label{fig:robot_space}
    \vspace{-8pt}
\end{figure}

\subsection{Dynamics Analysis}
The 3-section TDCR utilized in this study, as shown in Fig.~\ref{fig:robot_space}(a), is driven by 9 motors, with each section articulated by a trio of actuators. To protect the compliant backbone from excessive axial compression or tension, we impose a zero-net-extension constraint and a maximum speed limit on the motor velocity commands. Let \(\mathcal{S}_k = \{3k-2, 3k-1, 3k\}\) denotes the set of motor indices for the \(k\)-th section, then for all \(i \in \mathcal{S}_k\) and \(k \in \{1, 2, 3\}\), the actual command \(u^i\) for the \(i\)-th motor is calculated by
\begin{equation}
u^i = \text{clip}\left(a^i - \frac{1}{3} \sum_{j \in \mathcal{S}_k} a^j, -u_{\max}, u_{\max}\right),
\label{eq:constraint}
\end{equation}
where \(\bm{a} \in \mathbb{R}^9\) represents a raw command and \(u_{max}\) is the maximum motor velocity.

In our model, we consider the actual command \(\bm{u}\) as the input and the tip pose \(\mathbf{T}\) as the output. Intermediate states, including motor response, effective tendon lengths, and tension distributions, are collectively categorized as variables in the configuration space. Consequently, our model establishes a relationship from the actuation space to the task space. In fact, it is composed of two distinct dynamic mappings:
\begin{itemize}
    \item \textbf{From the actuation to configuration space.} The control command \(\bm{u}\) does not instantaneously equate to the actual motor velocity \(\bm{v}\) due to the inherent response latency of motors, which is strongly coupled with the motor torque induced by tendon tension and friction.
    \item \textbf{From the configuration to task space.} The viscoelastic deformation and transmission compliance make the motor rotation length \(\bm{l}\) and velocity \(\bm{v}\) not linearly transmitted to the effective tendon length and extension velocity, while the tip pose \(\mathbf{T}\) is a collective result of the effective tendon lengths, motor torques, gravity, and the friction along the routing paths.
\end{itemize}

To capture the above two mappings, we utilize onboard sensors at the motor side to measure a subset of configuration-space variables, specifically the motor rotation length \(\bm{l}\) and velocity \(\bm{v}\). Additionally, for high-precision trajectory tracking, a motion capture system is employed to provide ground-truth measurements of pose \(\mathbf{T}\).

\subsection{Dynamics Modeling}
As shown in Fig.~\ref{fig:robot_space}(b), we consolidate the physical system and its sensing modalities into a union dynamics model. The control input is defined as the actual motor velocity commands \(\bm{u}\), while the observation vector is given by
\begin{equation}
    \bm{o} = (\bm{l}, \bm{v}, \bm{p}, \bm{\phi}),
\end{equation}
where we parameterize the task-space pose $\mathbf{T}$ as the position vector $\bm{p} \in \mathbb{R}^3$ and the rotation vector $\bm{\phi} \in \mathbb{R}^3$.

\begin{figure}[t]
    \vspace{5pt}
    \centering
    \includegraphics[width=3.35in]{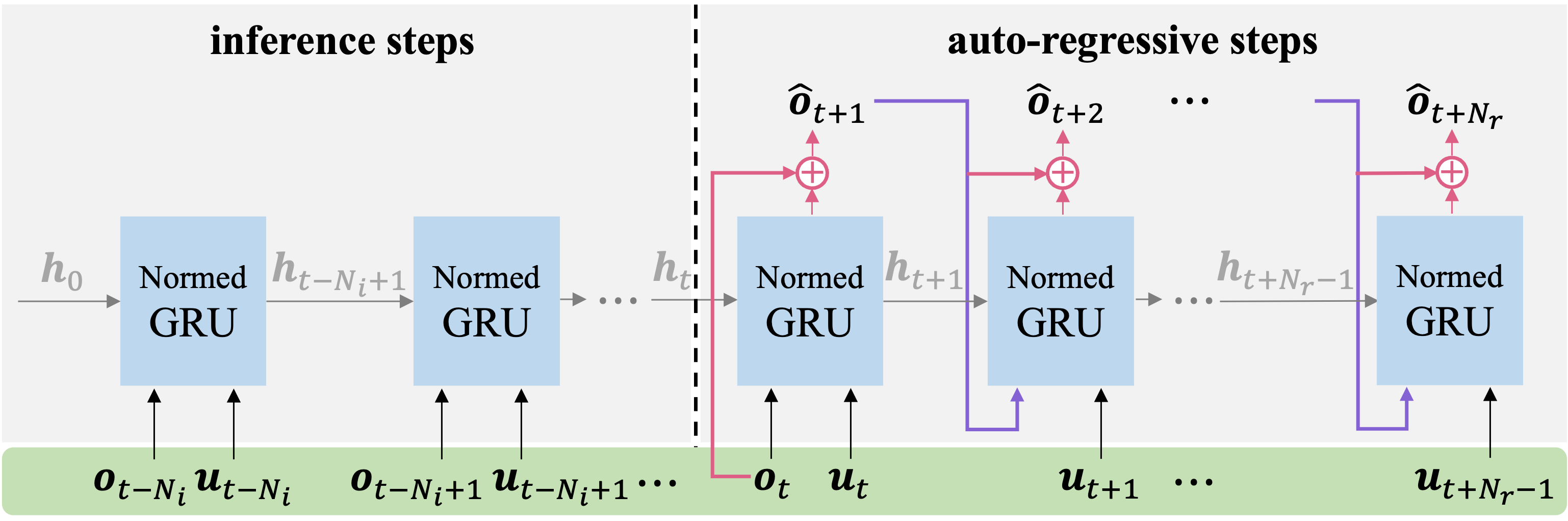}
    \vspace{-5pt}
    \caption{Training pipeline of the dynamics model. During inference steps, the hidden state \(\bm{h}\) is updated recurrently to incorporate historical context. During auto-regressive steps, the model performs multi-step rollouts where both the hidden state and the predicted observations (purple arrows) are fed back into the pipeline, incorporating residual prediction (pink arrows).}
    \label{fig:multi_res_gru}
    \vspace{-8pt}
\end{figure}

We employ GRU as the backbone of the dynamics model to capture the temporal dependencies within the system's response. The model maintains a hidden state \(\bm{h}\), which evolves at each time step to encode the historical context of the robot state. To improve prediction precision over long horizons and capture the underlying physical continuity, we formulate the observation prediction in a residual manner. Specifically, the dynamics model is formulated as  
\begin{equation}
    \label{eq:dynamics}
    \left\{
    \begin{aligned}
        \bm{h}_{t+1} &= f_{\theta}(\bm{h}_t, \bm{o}_t, \bm{u}_t), \\
        \Delta \hat{\bm{o}}_t &= g_{\psi}(\bm{h}_{t+1}), \\
        \hat{\bm{o}}_{t+1} &= \bm{o}_t + \Delta \hat{\bm{o}}_t.
    \end{aligned}
    \right.
\end{equation}

As illustrated in Fig.~\ref{fig:multi_res_gru}, the training pipeline is partitioned into two phases.
For the inference phase, we initialize the hidden state \(\bm{h}_0\) as a zero vector to serve as a common starting point. 
Subsequently, the model undergoes a warm-up process starting from step \(t-N_i\) to step \(t\). During this period, ground-truth observations and control inputs are sequentially fed into the GRU to refine the hidden state via the transition function \(f_{\theta}\). This procedure ensures that the latent representation \(\bm{h}_t\) sufficiently encodes the system's temporal context before performing future predictions.
In the auto-regressive phase, the model performs multi-step rollouts to simulate long-term dynamics. Leveraging multi-channel connectivity for better training, not only the updated hidden states but also the predicted observations are iteratively fed back as inputs for subsequent steps, together with the ground-truth control inputs.

The dynamics model is trained by minimizing the multi-step prediction error defined as
\begin{equation}
    \mathcal{L}_{\text{modeling}} = \sum_{i=1}^{N_r} \left\| \hat{\bm{o}}_{t+i} - \bm{o}_{t+i} \right\|^2,
    \label{eq:dynamics_loss}
\end{equation}
where \(N_r\) is the optimization horizon. 

Under this differentiable framework, the prediction loss in \eqref{eq:dynamics_loss} is backpropagated through both the recurrent hidden states and the iteratively predicted observations.

\section{Neural Control Policy}
In this section, we introduce a neural control policy that leverages latent temporal context, and is optimized through an differentiable pipeline. 

Traditional control of TDCRs typically employs variants of Jacobian-based kinematic schemes. Specifically, we consider three representative control laws listed below:
\begin{flalign}
&\text{(Feedback control)}&&\bm{u}=\mathbf{J}^{\dagger}\mathbf{K}\bm{e},&&\\
&\text{(Feedforward control)}&&\bm{u}=\mathbf{J}^{\dagger}\dot{\bm{r}},&&\\
&\text{(Hybrid control)}&&\bm{u}=\mathbf{J}^{\dagger}(\dot{\bm{r}}+\mathbf{K}\bm{e}),&&
\end{flalign}
where \(\bm{r}\) represents the reference point, \(\bm{e} = \bm{r} - \bm{o}\) is the tracking error, \(\mathbf{K}\) is a gain matrix, \(\mathbf{J}=\partial \bm{o} / \partial \bm{u}\) denotes the Jacobian, and \(\mathbf{J}^{\dagger}\) is the corresponding pseudo-inverse.

Although intuitive and widely adopted, these controllers suffer from fundamental theoretical limitations when applied to TDCR systems.
The primary deficit lies in the first-order Markovian assumption inherent in these formulations, which treats the robot as a quasi-static geometric entity rather than a higher-order dynamic system. 
In reality, the system is non-Markovian due to frictional hysteresis and viscoelasticity, making the mapping \(\bm{u} \to \bm{o}\) a path-dependent functional.
Consequently, the Jacobian \(\mathbf{J}\) is a multi-valued operator. Its value at any pose is not unique but depends on the internal state and movement history.

\begin{figure}[t]
    \vspace{5pt}
    \centering
    \includegraphics[width=2.4in]{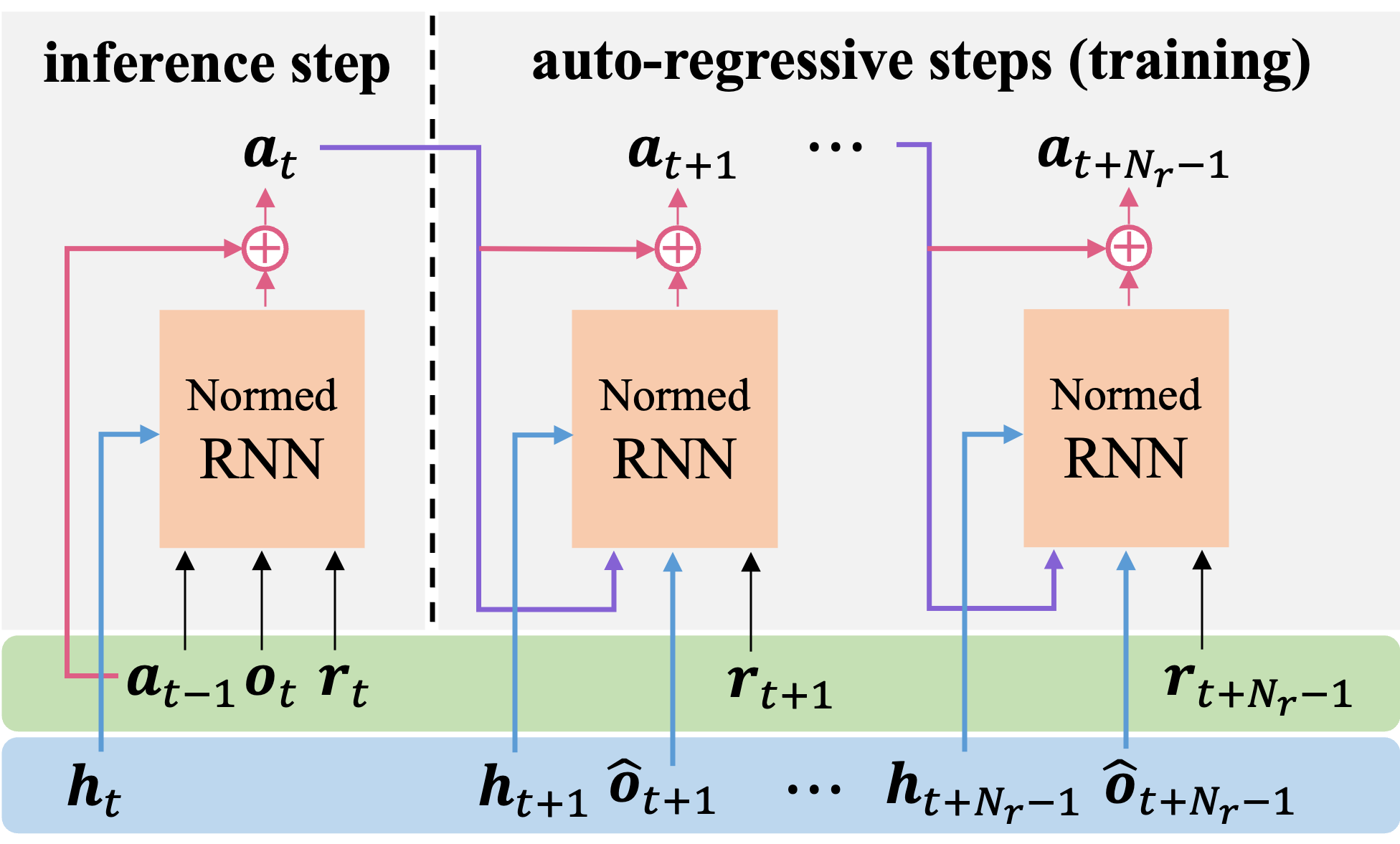}
    \vspace{-5pt}
    \caption{Training pipeline of the neural control policy. During auto-regressive steps, the policy recursively incorporates the hidden states and the predicted observations from the dynamics model (blue arrows), and generates residual actions (pink arrows) that are fed back into the pipeline (purple arrows).}
    \label{fig:policy}
    \vspace{-8pt}
\end{figure}

To bridge this gap, our policy incorporates the hidden state \(\bm{h}\) from the learned dynamics model. By reformulating the control problem in a latent space that accounts for historical context, the policy can effectively transforming a non-Markovian physical process into a well-conditioned tracking task. We employ an RNN as the backbone of our control policy. Unlike memory-less controllers, this architecture enables the policy to act as an implicit non-linear observer utilizing the historical context embedded in \(\bm{h}\). To improve training stability, we also formulate the action output in a residual manner. The neural control policy is given by
\begin{equation}
    \label{eq:policy}
    \left\{
    \begin{aligned}
        &\Delta \bm{a}_t = \pi_{\mu}(\bm{h}_t, \bm{a}_{t-1}, \bm{o}_t, \bm{r}_t), \\
        &\bm{a}_t = \bm{a}_{t-1} + \Delta \bm{a}_t,
    \end{aligned}
    \right.
\end{equation}
where \(\bm{r}_t\) represents the desired tip pose in future \(N_f\) steps, i.e.,
\begin{equation}
    \bm{r}_t = \left( \bm{p}_{t+1}^*, \bm{\phi}_{t+1}^*, \dots, \bm{p}_{t+N_f}^*, \bm{\phi}_{t+N_f}^* \right).
    \label{eq:reference}
\end{equation}
The multi-step reference provides the policy with preview capability, allowing it to proactively compensate for the time-lag and hysteresis inherent in TDCR dynamics. 
During training, we apply random shifts to the reference trajectories within the dataset to provide a broader distribution.

The training pipeline of the control policy is illustrated in Fig.~\ref{fig:policy}. 
During inference, the dynamics model recurrently evolves the hidden state \(\bm{h}\) to encapsulate the historical context of the TDCR. Subsequently, the policy recursively leverages the hidden states and the predicted observations from the dynamics model to generate control commands. These commands are then subjected to the zero-net-extension constraint and the maximum speed limit defined in \eqref{eq:constraint}, which are then fed back to actuate the dynamics model for the next auto-regressive step.

The control policy is optimized by minimizing the multi-step tracking error of the tip pose 
\begin{equation}
    \label{eq:policy_tracking_loss}
    \mathcal{L}_{\text{tracking}} = \sum_{i=1}^{N_r} \lambda^{i-1} (\left\| \hat{\bm{p}}_{t+i} - \bm{p}^*_{t+i} \right\|^2 + \left\| \hat{\bm{\phi}}_{t+i} - \bm{\phi}^*_{t+i} \right\|^2)
\end{equation}
and a control smoothness loss 
\begin{equation}
    \mathcal{L}_{\text{smooth}} = \sum_{i=1}^{N_r-1} \lambda^{i-1} \left\| \bm{a}_{t+i} - \bm{a}_{t+i-1} \right\|^2
    \label{eq:policy_smooth_loss}
\end{equation}
over the horizon \(N_r\), where \(\lambda\) is the discount factor to prioritize immediate tracking accuracy and ensure optimization stability over long horizons.

The fully differentiable control-dynamics chain allows the dynamics model to serve not merely as a predictor, but as a gradient bridge during training. 
Consequently, the policy network can directly perceive how its actions propagate to influence future observations, enabling precise long-horizon optimization of the tracking performance.
Notably, this architecture is highly efficient for real-time implementation. Since the policy only requires a single forward pass of \(\pi_\mu\) and \(f_\theta\) within several milliseconds, it bypasses the need for computationally expensive full-horizon trajectory optimization while preserving long-term temporal awareness.

\section{Experiments}
\subsection{Setup}
\subsubsection{Hardware platform} 
The hardware platform comprises a motion capture system, motors, and the continuum robot. The robot is actuated by 9 motors integrated in the base, driving the tendons to modulate its configuration. The motors are interfaced with a PLC, receiving control commands from the host computer and monitoring the states of the motors. For tip pose feedback, the motion capture system comprising 5 infrared cameras tracks the robot. While operating at 120~Hz, the data are downsampled to 50~Hz to synchronize with our control frequency.

\subsubsection{Dataset} 
To construct the dataset \(\mathcal{D} = \{ \left( \bm{o}_t, \bm{u}_t, \bm{r}_t \right) \}\), a baseline feedforward-feedback controller was employed to track 16 distinct trajectories. Various controller parameters were utilized to sufficiently excite the system's underlying dynamics. These trajectories included both structured and random paths, augmented with control or reference noise at 1~Hz, 5~Hz, and stochastic frequencies to capture the system's step and impulse responses. Data was collected over six separate days with the system re-zeroed at the start of each session, yielding a comprehensive dataset of 923,059 steps (approximately 5.1 hours). To rigorously evaluate the model, the data was partitioned into four subsets:
\begin{itemize}
    \item \(\mathcal{D}_{train}\): Consisting of 707,732 steps from the initial five-day collection, excluding any ``T'' trajectories, used for model training.
    \item \(\mathcal{D}_{test}\): Consisting of 157,820 steps from the initial five-day collection, excluding any ``T'' trajectories, used for validation.
    \item \(\mathcal{D}_{traj}\): Consisting of 29,532 steps of only ``T'' trajectories used to evaluate generalization across different motion patterns.
    \item \(\mathcal{D}_{date}\):  Consisting of 27,975 steps collected on the separate day used to assess robustness against temporal dynamics drift.
\end{itemize}

\begin{figure}[t]
    \vspace{5pt}
    \centering
    \includegraphics[width=3.2in]{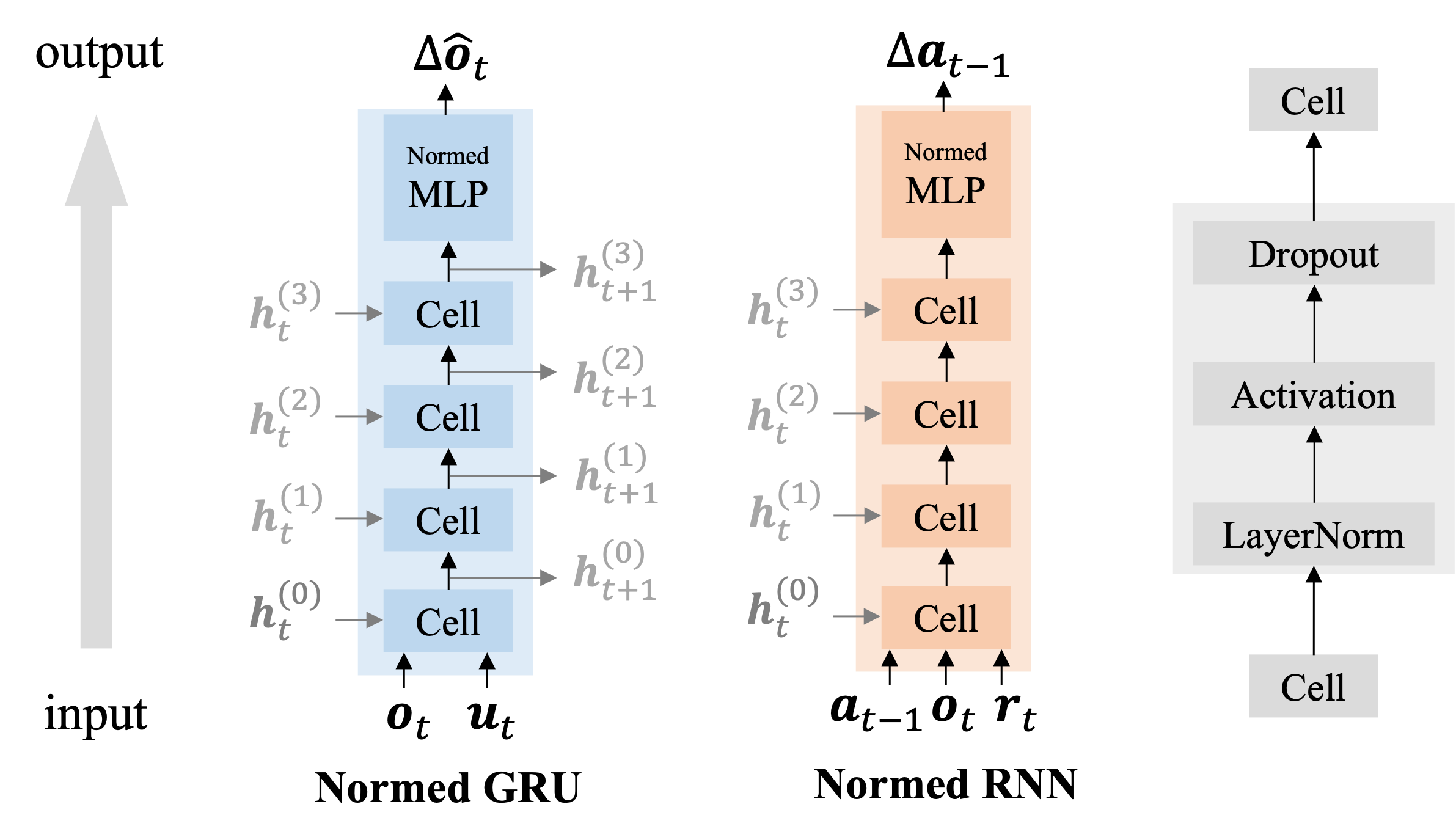}
    \vspace{-8pt}
    \caption{Architecture of the 4-layer RNNs used in our model. LayerNorm~\cite{ba2016layernormalization}, activation, and dropout~\cite{nitish2014dropout} are applied after each hidden layer.}
    \label{fig:network}
    \vspace{-8pt}
\end{figure}

\subsubsection{Networks and parameters} 
The proposed network architecture is illustrated in Fig.~\ref{fig:network}. We employ a 4-layer RNN structure to capture complex, high-order dynamics, with LayerNorm~\cite{ba2016layernormalization} integrated after each hidden layer to enhance training stability. 
The model is configured with a hidden state dimension of 1024, and a dropout rate of 0.3 to prevent overfitting. During the dynamics training phase, we set the inference horizon and the auto-regressive length to \(N_i = 50\) and \(N_r=50\), respectively. For control policy optimization, the auto-regressive length is extended to \(N_r=250\) to ensure the convergence of trajectory tracking, and the discount factor is set to \(\lambda = 0.98\) balance immediate precision with long-term predictive stability.

To mitigate scale discrepancies across modalities, all inputs are standardized to a unit normal distribution based on the whole dataset's mean and standard deviation before entering the network, with model outputs subsequently denormalized to retrieve their original physical units.

\begin{table}[t]
    \centering
    \vspace{10pt}
    \caption{Average First-Step Prediction Errors across Datasets (mm / \textdegree)}
    \label{tab:dynamics_dataset}
    \renewcommand{\arraystretch}{1.2}
    \begin{tabular}{ccccc}
        \toprule
        \multicolumn{2}{c}{Training Set} & \(\mathcal{D}_{test}\) & \(\mathcal{D}_{traj}\) & \(\mathcal{D}_{date}\) \\
        \midrule
        \(\mathcal{D}_{train}\) & 1.27 / 1.33\textdegree & \textbf{1.43} / \textbf{1.40\textdegree} & \textbf{1.48} / \textbf{0.84\textdegree} & \textbf{1.32} / \textbf{1.23\textdegree} \\
        \(\mathcal{D}_{small}\) & 1.22 / 1.31\textdegree & 1.58 / 1.64\textdegree & 1.86 / 1.12\textdegree & 1.57 / 1.39\textdegree \\
        \(\mathcal{D}_{2days}\) & 0.79 / 0.72\textdegree & 2.18 / 1.63\textdegree & 4.66 / 1.10\textdegree & 3.67 / 1.94\textdegree \\
        \(\mathcal{D}_{clean}\) & 0.90 / 0.80\textdegree & 2.29 / 2.03\textdegree & 2.28 / 2.66\textdegree & 1.67 / 1.39\textdegree \\
        \bottomrule
    \vspace{-20pt}
    \end{tabular}
\end{table}

\begin{figure}[t]
    \vspace{5pt}
    \centering
    \includegraphics[width=3.2in]{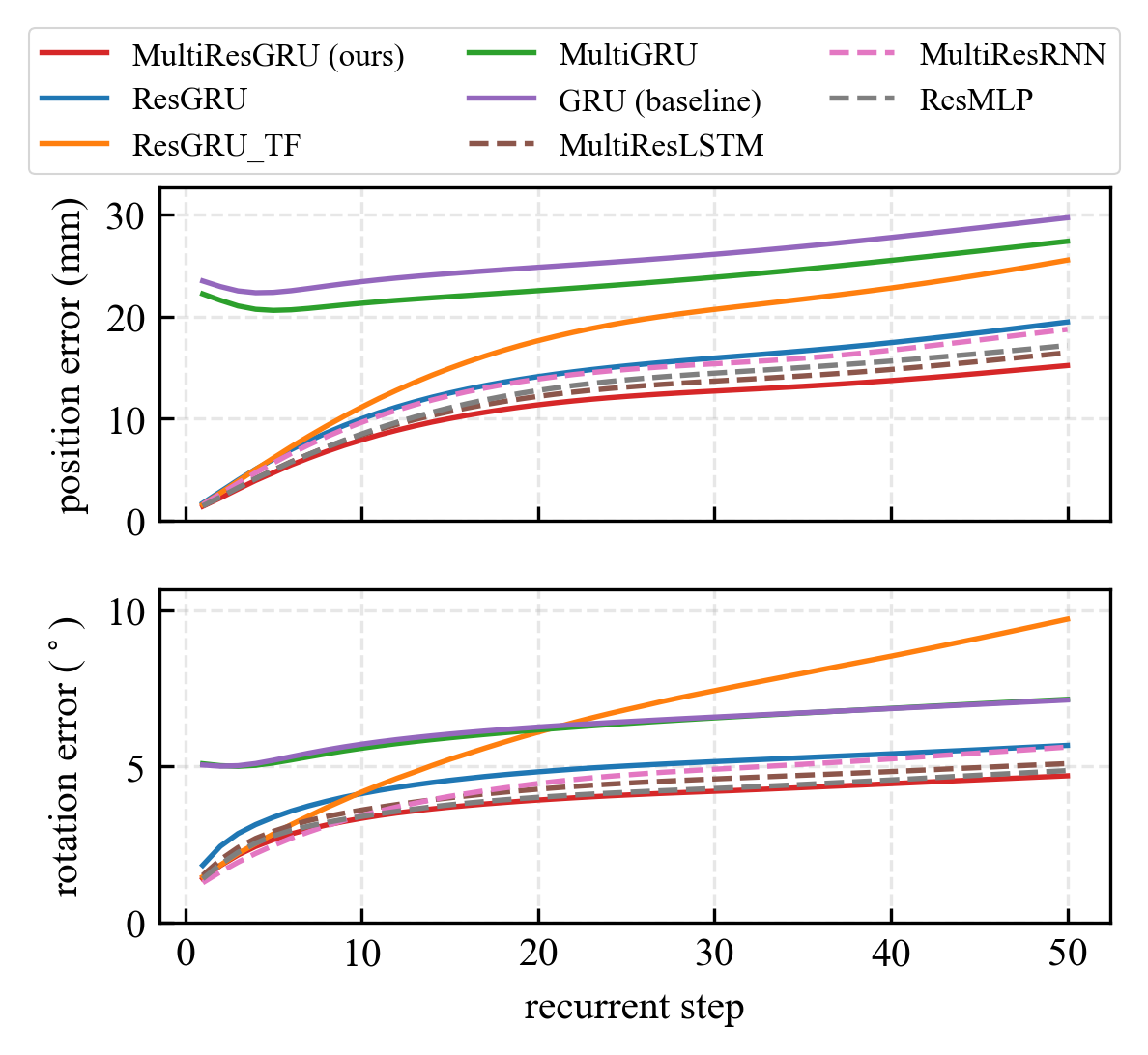}
    \vspace{-12pt}
    \caption{Average position and rotation errors of different model configurations relative to auto-regressive prediction steps on \(\mathcal{D}_{test}\). Our method (red solid line) achieves the lowest prediction error, while exhibiting the slowest error accumulation rate as the prediction horizon extends.}
    \label{fig:dynamics_step}
    \vspace{-8pt}
\end{figure}

\begin{figure*}[t]
    \vspace{5pt}
    \centering
    \includegraphics[width=6.8in]{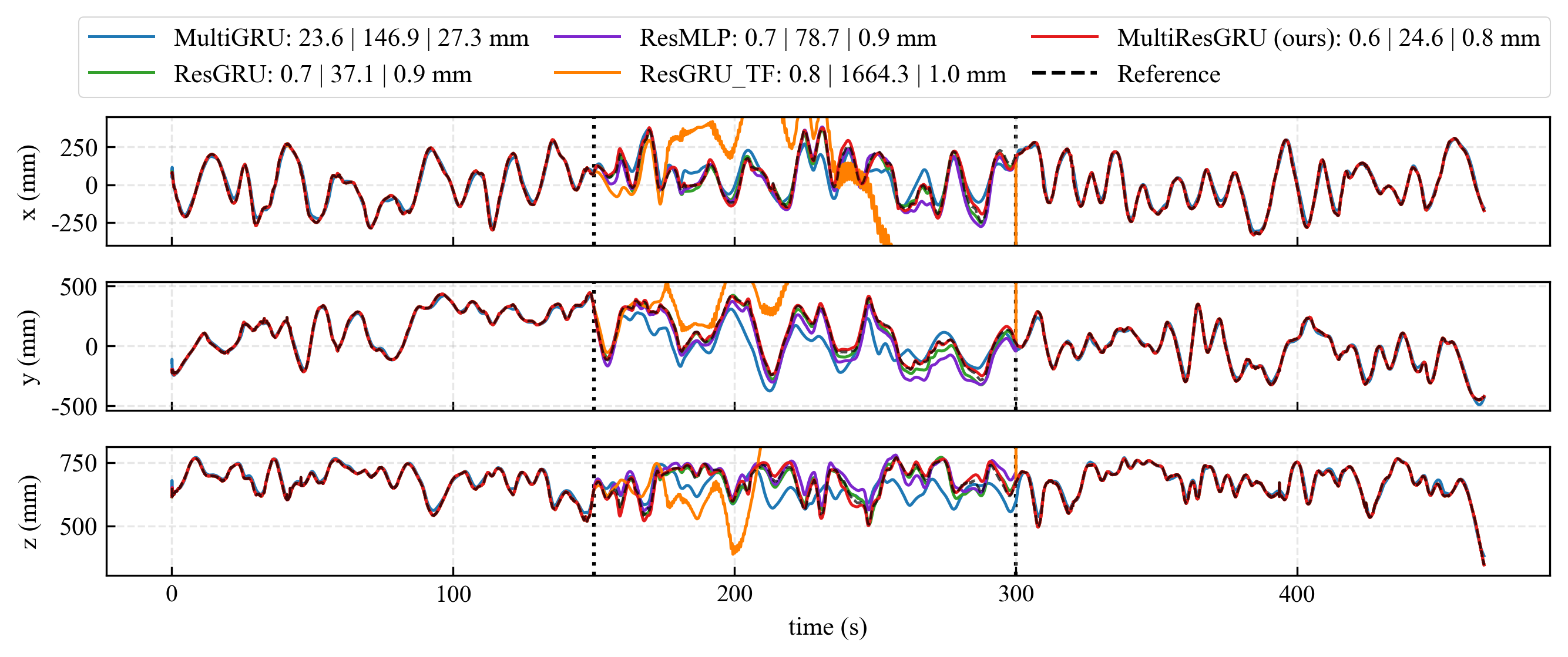}
    \vspace{-12pt}
    \caption{Position prediction performance of different model configurations across a long random trajectory. The evaluation consists of three phases: one-step prediction from 0 to 150~s, auto-regressive prediction from 150 to 300~s, and a reversion to one-step prediction after 300~s. The legend indicates the average position errors for each configuration across all phases.}
    \label{fig:dynamics_traj}
    \vspace{-8pt}
\end{figure*}

\subsection{Performance of the Dynamics Model}
In this section, we evaluate the prediction performance of the proposed dynamics model. Specifically, we aim to investigate
(a) the \textbf{generalization and adaptation} capability of the model to unseen trajectories and dynamics shifts driven by dataset scale and diversity,
(b) the \textbf{long-term precision and stability} of the model during auto-regressive prediction enhanced by residual and bidirectional multi-channel structures,
and (c) the \textbf{comparative performance} of our proposed architecture against alternative network structures.

\subsubsection{Generalization and adaptation} 
To evaluate the impact of training data on model accuracy and generalization, we construct three subsets of \(\mathcal{D}_{train}\) as follows:
\begin{itemize}
    \item \(\mathcal{D}_{small}\): Consisting of 339,338 steps randomly sampled from \(\mathcal{D}_{train}\) to evaluate the impact of data scale.
    \item \(\mathcal{D}_{2days}\): Consisting of 336,747 steps comprising trajectories of two days to assess limited temporal coverage.
    \item \(\mathcal{D}_{clean}\): Consisting of 333,031 steps comprising only noise-free trajectories to investigate the impact of limited dynamics patterns.
\end{itemize}

To ensure a consistent total iteration steps, our model is trained for 100 epochs on \(\mathcal{D}_{train}\) and 200 epochs on each smaller subsets (\(\mathcal{D}_{small}\), \(\mathcal{D}_{2days}\), and \(\mathcal{D}_{clean}\)). We evaluate the average first-step prediction errors of the tip position (mm) and orientation (\textdegree), across both training and unseen test sets (\(\mathcal{D}_{small}\), \(\mathcal{D}_{2days}\), and \(\mathcal{D}_{clean}\)), as shown in Table~\ref{tab:dynamics_dataset}.

The model trained on \(\mathcal{D}_{train}\) exhibits consistent performance across all test sets, demonstrating its ability to generalize across diverse trajectories and mitigate temporal dynamics drift. While the model trained on \(\mathcal{D}_{small}\) achieves comparable training accuracy to \(\mathcal{D}_{train}\), its reduced performance on test sets underscores the importance of data scale for generalization. Furthermore, although models trained on \(\mathcal{D}_{2days}\) and \(\mathcal{D}_{clean}\) perform well on their respective training sets, their lower accuracy on other test sets suggests that both data diversity and high-frequency response information (contained in control noise) are crucial for model robustness.

\subsubsection{Comparison and long-term precision} 
We also compare our proposed method against several baseline architectures to validate each component's contribution, including
\begin{itemize}
    \item \textbf{MultiResGRU (ours):} Our method that leverage bidirectional multi-connectivity both in the hidden states and the predicted observations with residual prediction.
    \item \textbf{ResGRU:} A variant that maintains auto-regressive connectivity of the predicted observations in the forward pass while detaching them from the gradient computational graph during backpropagation, restricting gradient flow solely through the hidden state path.
    \item \textbf{ResGRU\_TF}: A configuration trained with teacher forcing, where both forward and backward connectivity through observations are severed by using ground-truth observations as inputs, leaving hidden states as the only temporal link.
    \item \textbf{MultiGRU:} A version without residual prediction, where the network outputs the absolute values of the predicted observations directly.
    \item \textbf{GRU (baseline)}: A standard GRU architecture lacking both bidirectional multi-channel connectivity and residual prediction.
    \item \textbf{MultiResLSTM / MultiResRNN}: Architectures employing Long Short-Term Memory (LSTM) or standard RNN cells as the backbone instead of GRU cells.  
    \item \textbf{ResMLP}: A Multi-Layer Perceptron (MLP) based model that takes a concatenated window of \(N_i\) historical observations and actions as input, updating the input buffer recursively with its own predictions.
\end{itemize}

\begin{figure*}[t]
    \vspace{5pt}
    \centering
    \includegraphics[width=6.7in]{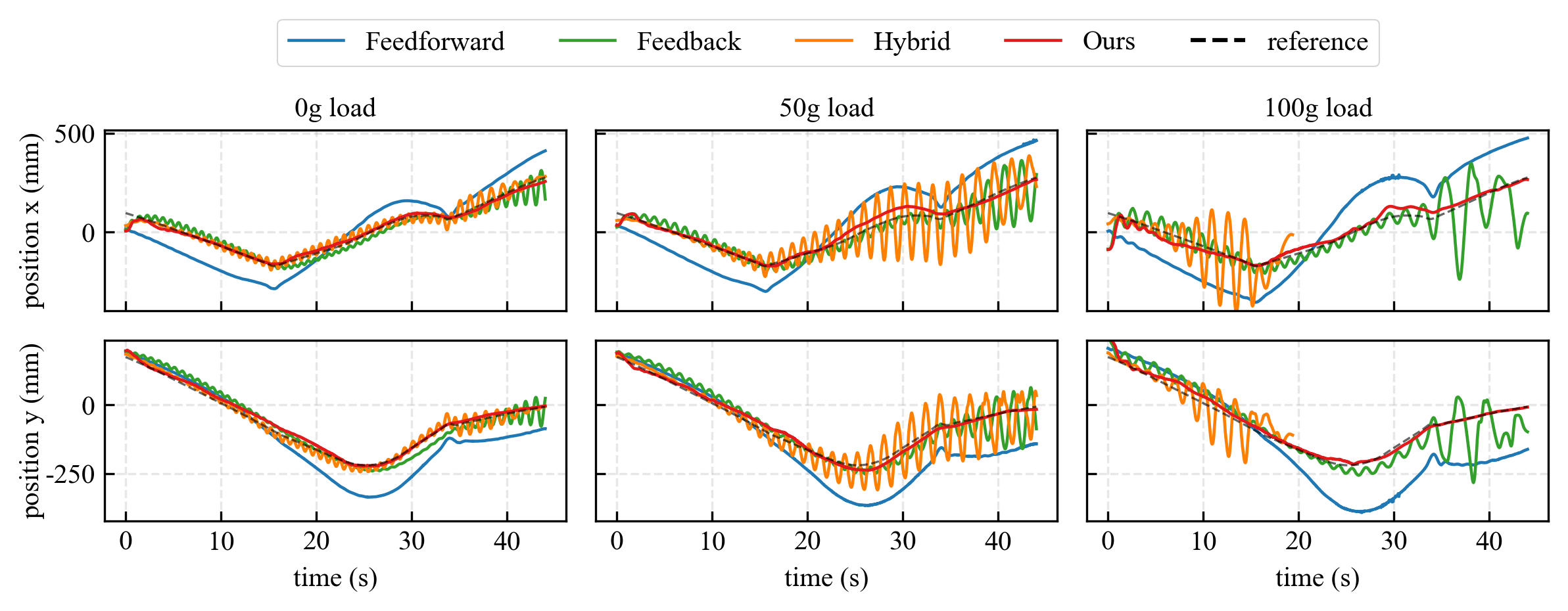}
    \vspace{-12pt}
    \caption{Tracking performance under varying payload disturbances (0g, 50g, and 100g). The baseline \textit{Feedback} and \textit{Hybrid} controllers exhibit self-excited oscillations as the load increases, while the \textit{Feedforward} method suffers from substantial steady-state error. In contrast, our neural control policy demonstrates superior robustness and maintains accurate tracking. The 100g trial for the \textit{Hybrid} controller was prematurely terminated to prevent hardware damage from divergent instability.}
    \label{fig:policy_traj}
    \vspace{-8pt}
\end{figure*}

As shown in Fig.~\ref{fig:dynamics_step}, models without residual prediction (MultiGRU and GRU baseline) exhibit initial errors exceeding 20~mm and 5\(^\circ\), whereas other configurations start around 1~mm and 1\(^\circ\), highlighting the critical role of residual learning in mitigating prediction errors.
While ResGRU\_TF starts accurately, it suffers from rapid error accumulation. In contrast, the slower error growth in ResGRU demonstrates that feeding predicted observations back into the model effectively suppresses compounding errors.
Our MultiResGRU further improves multi-step precision by enabling gradient flow through this observation path.
Among the backbones, MultiResGRU performs best, likely because its unified update gate provides a more suitable inductive bias for the continuous dynamics.

A more challenging evaluation of long-term auto-regression is presented in Fig.~\ref{fig:dynamics_traj}, where the models undergo a continuous 2.5-minute auto-regressive prediction phase.
During this stage, the models receive only control actions without any ground-truth observation feedback.
MultiResGRU (ours) achieves a mean prediction error of only 24.6~mm, and maintains high fidelity to the trajectory details.
In contrast, ResGRU with only forward observation paths and ResMLP without hidden state connectivity fail to reconstruct details at trajectory extrema.
Meanwhile, ResGRU\_TF, which links time steps only via hidden states, diverges rapidly during multi-step prediction.
These results underscore that our bidirectional multi-channel connectivity is instrumental in enhancing multi-step precision and stability. 
Furthermore, MultiGRU avoids total divergence but suffers from significant prediction bias without residual learning.

\subsection{Performance of the Neural Control Policy}
In this section, we evaluate the tracking performance of the proposed control policy. Specifically, we aim to investigate (a) the \textbf{tracking accuracy} and (b) the \textbf{robustness} to disturbances of our neural control policy in comparison with Jacobian-based controllers.

\begin{table}[t]
    \centering
    \vspace{10pt}
    \caption{Average Tracking Errors of Different Controllers (mm / \textdegree)}
    \label{tab:policy_speed}
    \renewcommand{\arraystretch}{1.2}
    \begin{tabular}{ccccc}
        \toprule
        Speed & \textit{Ours} & \textit{Feedforward} & \textit{Feedback} & \textit{Hybrid} \\
        \midrule
        1.0x & 25.3 / \textbf{7.2\textdegree} & 159.1 / 16.4\textdegree & 46.6 / 26.5\textdegree & \textbf{24.5} / 23.1\textdegree \\
        1.7x & \textbf{30.4} / \textbf{8.2\textdegree} & 159.0 / 16.5\textdegree & 62.8 / 23.7\textdegree & 33.0 / 21.9\textdegree \\
        2.5x & \textbf{36.1} / \textbf{10.0\textdegree} & 159.1 / 16.4\textdegree & 79.6 / 22.8\textdegree & 43.6 / 20.9\textdegree \\
        \bottomrule
    \vspace{-20pt}
    \end{tabular}
\end{table}

\subsubsection{Tracking accuracy} 
To evaluate the controllers, we conduct tracking experiments on five distinct trajectories within the $xy$-plane at a constant height, including ``R'', ``O'', ``B'', ``T'' shapes, and a straight line ``--''.

The average tracking errors are summarized in Table~\ref{tab:policy_speed}. The results show that the \textit{Feedforward} controller exhibits the largest errors, primarily due to its inability to compensate for accumulated drifts as an open-loop method. While the \textit{Feedback} controller improves performance, it remains inferior to the \textit{Hybrid} method, which benefits from combining nominal dynamics with reactive corrections. 
Notably, at lower speeds (1.0x, around 23~mm/s), our neural control policy achieves position accuracy comparable to the \textit{Hybrid} controller while providing a significant reduction in rotation error. Furthermore, as the execution speed increases (1.7x to 2.5x, around 58~mm/s), our method presents a widening performance advantage. This demonstrates that our neural policy more effectively captures the system's high-frequency dynamics and nonlinearities, which are often neglected in kinematic controller.

\subsubsection{Robustness} 
To evaluate the robustness of the controllers under external disturbances, we conduct tracking experiments with payloads of 50g and 100g attached to the robot's end-effector, which are not seen in training data. As illustrated in Fig.~\ref{fig:policy_traj}, both \textit{Feedback} and \textit{Hybrid} controllers experience significant self-excited oscillations, which become increasingly pronounced as the payload increases. While the \textit{Feedforward} method avoids such oscillations, its tracking deviation grows substantially with the added load. In contrast, our proposed neural control policy maintains stable and effective tracking performance regardless of the payload conditions, demonstrating superior robustness.

\section{Conclusion}
This work presents a differentiable recurrent framework for the high-fidelity modeling and robust control of TDCRs. By integrating bidirectional gradient flow with residual learning and leveraging diverse datasets, our dynamics model successfully captures path-dependent nonlinearities and adapts to temporal dynamics shifts, all while effectively suppressing compounding errors during long-horizon predictions. This allows the system's non-Markovian physics to be implicitly internalized within a latent space, bypassing the need for explicit parameter identification. Hardware experiments confirm that the resulting neural policy achieves centimeter-level tracking precision and eliminates the self-excited oscillations prevalent in Jacobian-based schemes.


\bibliographystyle{IEEEtran}
\bibliography{ref}

@article{burgner2015survey,
  title={Continuum robots for medical applications: A survey},
  author={Burgner-Kahrs, Jessica and Rucker, D Caleb and Choset, Howie},
  journal={IEEE Transactions on Robotics},
  volume={31},
  number={6},
  pages={1261--1280},
  year={2015},
  publisher={IEEE}
}

@article{webster2010design,
  title={Design and kinematic modeling of constant curvature continuum robots: A review},
  author={Webster III, Robert J and Jones, Bryan A},
  journal={The International Journal of Robotics Research},
  volume={29},
  number={13},
  pages={1661-1683},
  year={2010},
  publisher={SAGE Publications Sage UK: London, England}
}

@article{hu2025contact,
  title={Contact Force Estimation of Continuum Robots without Embedded Sensors: A Review},
  author={Hu, An and Sun, Yu},
  journal={Advanced Intelligent Systems},
  pages={e202500786},
  year={2025},
  publisher={Wiley Online Library}
}

@article{till2019real,
  title={Real-time dynamics of soft and continuum robots based on {C}osserat rod models},
  author={Till, John and Aloi, Vincent and Rucker, Caleb},
  journal={The International Journal of Robotics Research},
  volume={38},
  number={6},
  pages={723--746},
  year={2019},
  publisher={SAGE Publications Sage UK: London, England}
}

@article{rucker2011statics,
  title={Statics and dynamics of continuum robots with general tendon routing and external loading},
  author={Rucker, D Caleb and Webster III, Robert J},
  journal={IEEE Transactions on Robotics},
  volume={27},
  number={6},
  pages={1033--1044},
  year={2011},
  publisher={IEEE}
}

@article{liu2025data,
  title={Data-driven methods for sensing, modeling and control of soft continuum robot: A review},
  author={Liu, Jiaqi and Duo, Youning and Chen, Xingyu and Zuo, Zonghao and Liu, Yuchen and Wen, Li},
  journal={IEEE/ASME Transactions on Mechatronics},
  year={2025},
  publisher={IEEE}
}

@article{wang2022control,
  title={Control strategies for soft robot systems},
  author={Wang, Jue and Chortos, Alex},
  journal={Advanced Intelligent Systems},
  volume={4},
  number={5},
  pages={2100165},
  year={2022},
  publisher={Wiley Online Library}
}

@article{campisano2021closed,
  title={Closed-loop control of soft continuum manipulators under tip follower actuation},
  author={Campisano, Federico and Cal{\'o}, Simone and Remirez, Andria A and Chandler, James H and Obstein, Keith L and Webster III, Robert J and Valdastri, Pietro},
  journal={The International Journal of Robotics Research},
  volume={40},
  number={6-7},
  pages={923--938},
  year={2021},
  publisher={SAGE Publications Sage UK: London, England}
}

@inproceedings{satheeshbabu2019open,
  title={Open loop position control of soft continuum arm using deep reinforcement learning},
  author={Satheeshbabu, Sreeshankar and Uppalapati, Naveen Kumar and Chowdhary, Girish and Krishnan, Girish},
  booktitle={2019 International Conference on Robotics and Automation (ICRA)},
  pages={5133--5139},
  year={2019},
  organization={IEEE}
}

@inproceedings{rucker2011computing,
  title={Computing jacobians and compliance matrices for externally loaded continuum robots},
  author={Rucker, D Caleb and Webster, Robert J},
  booktitle={2011 IEEE international conference on robotics and automation},
  pages={945--950},
  year={2011},
  organization={IEEE}
}

@article{almanzor2023static,
  title={Static shape control of soft continuum robots using deep visual inverse kinematic models},
  author={Almanzor, Elijah and Ye, Fan and Shi, Jialei and Thuruthel, Thomas George and Wurdemann, Helge A and Iida, Fumiya},
  journal={IEEE Transactions on Robotics},
  volume={39},
  number={4},
  pages={2973--2988},
  year={2023},
  publisher={IEEE}
}

@inproceedings{grassmann2018learning,
  title={Learning the forward and inverse kinematics of a 6-DOF concentric tube continuum robot in SE (3)},
  author={Grassmann, Reinhard and Modes, Vincent and Burgner-Kahrs, Jessica},
  booktitle={2018 IEEE/RSJ International Conference on Intelligent Robots and Systems (IROS)},
  pages={5125--5132},
  year={2018},
  organization={IEEE}
}

@article{thuruthel2017learning,
  title={Learning dynamic models for open loop predictive control of soft robotic manipulators},
  author={Thuruthel, Thomas George and Falotico, Egidio and Renda, Federico and Laschi, Cecilia},
  journal={Bioinspiration \& biomimetics},
  volume={12},
  number={6},
  pages={066003},
  year={2017},
  publisher={IOP Publishing}
}

@article{qiu2025actuator,
  title={An actuator space optimal kinematic path tracking framework for tendon-driven continuum robots: Theory, algorithm and validation},
  author={Qiu, Ke and Zhang, Hongye and Zhang, Jingyu and Xiong, Rong and Lu, Haojian and Wang, Yue},
  journal={The International Journal of Robotics Research},
  volume={44},
  number={6},
  pages={1006--1034},
  year={2025},
  publisher={Sage Publications Sage UK: London, England}
}

@inproceedings{kazemipour2022adaptive,
  title={Adaptive dynamic sliding mode control of soft continuum manipulators},
  author={Kazemipour, Amirhossein and Fischer, Oliver and Toshimitsu, Yasunori and Wong, Ki Wan and Katzschmann, Robert K},
  booktitle={2022 International Conference on Robotics and Automation (ICRA)},
  pages={3259--3265},
  year={2022},
  organization={IEEE}
}

@article{george2017learning,
  title={Learning closed loop kinematic controllers for continuum manipulators in unstructured environments},
  author={George Thuruthel, Thomas and Falotico, Egidio and Manti, Mariangela and Pratesi, Andrea and Cianchetti, Matteo and Laschi, Cecilia},
  journal={Soft robotics},
  volume={4},
  number={3},
  pages={285--296},
  year={2017},
  publisher={Mary Ann Liebert, Inc. 140 Huguenot Street, 3rd Floor New Rochelle, NY 10801 USA}
}

@book{lynch2017modern,
  title={Modern robotics},
  author={Lynch, Kevin M and Park, Frank C},
  year={2017},
  publisher={Cambridge University Press}
}

@inproceedings{gonthina2020mechanics,
  author={Gonthina, Phanideep S. and Wooten, Michael B. and Godage, Isuru S. and Walker, Ian D.},
  booktitle={2020 IEEE International Conference on Robotics and Automation (ICRA)}, 
  title={Mechanics for Tendon Actuated Multisection Continuum Arms}, 
  year={2020},
  volume={},
  number={},
  pages={3896-3902}
}

@article{della2020model,
  title={Model-based dynamic feedback control of a planar soft robot: trajectory tracking and interaction with the environment},
  author={Della Santina, Cosimo and Katzschmann, Robert K and Bicchi, Antonio and Rus, Daniela},
  journal={The International Journal of Robotics Research},
  volume={39},
  number={4},
  pages={490--513},
  year={2020},
  publisher={SAGE Publications Sage UK: London, England}
}

@article{khadem2020autonomous,
  title={Autonomous steering of concentric tube robots via nonlinear model predictive control},
  author={Khadem, Mohsen and O'Neill, John and Mitros, Zisos and Da Cruz, Lyndon and Bergeles, Christos},
  journal={IEEE Transactions on Robotics},
  volume={36},
  number={5},
  pages={1595--1602},
  year={2020},
  publisher={IEEE}
}

@inproceedings{spinelli2022unified,
  title={A unified and modular model predictive control framework for soft continuum manipulators under internal and external constraints},
  author={Spinelli, Filippo A and Katzschmann, Robert K},
  booktitle={2022 IEEE/RSJ International Conference on Intelligent Robots and Systems (IROS)},
  pages={9393--9400},
  year={2022},
  organization={IEEE}
}

@article{amanov2021tendon,
  title={Tendon-driven continuum robots with extensible sections—A model-based evaluation of path-following motions},
  author={Amanov, Ernar and Nguyen, Thien-Dang and Burgner-Kahrs, Jessica},
  journal={The International Journal of Robotics Research},
  volume={40},
  number={1},
  pages={7--23},
  year={2021},
  publisher={SAGE Publications Sage UK: London, England}
}

@article{boyer2020dynamics,
  title={Dynamics of continuum and soft robots: A strain parameterization based approach},
  author={Boyer, Frederic and Lebastard, Vincent and Candelier, Fabien and Renda, Federico},
  journal={IEEE Transactions on Robotics},
  volume={37},
  number={3},
  pages={847--863},
  year={2020},
  publisher={IEEE}
}

@article{renda2020geometric,
  title={A geometric variable-strain approach for static modeling of soft manipulators with tendon and fluidic actuation},
  author={Renda, Federico and Armanini, Costanza and Lebastard, Vincent and Candelier, Fabien and Boyer, Frederic},
  journal={IEEE Robotics and Automation Letters},
  volume={5},
  number={3},
  pages={4006--4013},
  year={2020},
  publisher={IEEE}
}

@INPROCEEDINGS{wen2017tdhand,
  author={Wen, Ligang and Li, Yongyao and Cong, Ming and Lang, Haoxiang and Du, Yu},
  booktitle={2017 IEEE International Conference on Industrial Technology (ICIT)}, 
  title={Design and optimization of a tendon-driven robotic hand}, 
  year={2017},
  volume={},
  number={},
  pages={767-772},
  doi={10.1109/ICIT.2017.7915456}
}

@article{zhang2024copilot,
    author = {Jingyu Zhang and Lilu Liu and Pingyu Xiang and Qin Fang and Xiuping Nie and Honghai Ma and Jian Hu and Rong Xiong and Yue Wang and Haojian Lu},
    title = {AI Co-Pilot Bronchoscope Robot},
    journal = {Nature Communications},
    year = {2024},
    volume={15},
    number={241},
}

@article{zhang2023hysteresis,
    author = {Yue Zhang and Lijin Fang and Tangzhong Song and Ming Zhang},
    title ={Model-free adaptive control based on prescribed performance and time delay estimation for robotic manipulators subject to backlash hysteresis},
    journal = {Proceedings of the Institution of Mechanical Engineers, Part C: Journal of Mechanical Engineering Science},
    volume = {237},
    number = {23},
    pages = {5674-5691},
    year = {2023},
    doi = {10.1177/09544062231160350},
}

@article{jennifer2015elastomers,
    author = {Jennifer C. Case and Edward L. White and Rebecca K. Kramer},
    title ={Soft Material Characterization for Robotic Applications},
    journal = {Soft Robotics},
    volume = {2},
    number = {2},
    pages = {80-87},
    year = {2015},
    doi = {10.1089/soro.2015.0002},
}

@ARTICLE{hendrik2024rnn,
  author={Schäfke, Hendrik and Habich, Tim-Lukas and Muhmann, Christian and Ehlers, Simon F. G. and Seel, Thomas and Schappler, Moritz},
  journal={IEEE Robotics and Automation Letters}, 
  title={Learning-Based Nonlinear Model Predictive Control of Articulated Soft Robots Using Recurrent Neural Networks}, 
  year={2024},
  volume={9},
  number={12},
  pages={11609-11616},
  doi={10.1109/LRA.2024.3495579}}

@ARTICLE{tim2026pinn,
  author={Habich, Tim-Lukas and Mohammad, Aran and Ehlers, Simon F. G. and Bensch, Martin and Seel, Thomas and Schappler, Moritz},
  journal={IEEE Transactions on Robotics}, 
  title={Generalizable and Fast Surrogates: Model Predictive Control of Articulated Soft Robots Using Physics-Informed Neural Networks}, 
  year={2026},
  volume={42},
  number={},
  pages={619-636},
  doi={10.1109/TRO.2025.3631818}
}

@ARTICLE{yu2026zhongshan,
  author={Yu, Peng and Liang, Zhenhan and Tan, Ning},
  journal={IEEE Transactions on Robotics}, 
  title={Data-Efficient and Predefined-Time Stable Control for Continuum Robots}, 
  year={2026},
  volume={42},
  number={},
  pages={382-399},
  doi={10.1109/TRO.2025.3644946}
}

@article{tang2026science,
    author = {Zhiqiang Tang  and Liying Tian  and Wenci Xin  and Qianqian Wang  and Daniela Rus  and Cecilia Laschi },
    title = {A general soft robotic controller inspired by neuronal structural and plastic synapses that adapts to diverse arms, tasks, and perturbations},
    journal = {Science Advances},
    volume = {12},
    number = {2},
    pages = {eaea3712},
    year = {2026},
}

@InProceedings{ross2011dagger,
  title = {A Reduction of Imitation Learning and Structured Prediction to No-Regret Online Learning},
  author = {Ross, Stephane and Gordon, Geoffrey and Bagnell, Drew},
  booktitle = {Proceedings of the Fourteenth International Conference on Artificial Intelligence and Statistics},
  pages = {627--635},
  year = {2011},
  editor = {Gordon, Geoffrey and Dunson, David and Dudík, Miroslav},
  volume = {15},
  series = {Proceedings of Machine Learning Research},
  address = {Fort Lauderdale, FL, USA},
  month = {11--13 Apr},
  publisher = {PMLR},
}

@InProceedings{haarnoja2018sac,
  title =  {Soft Actor-Critic: Off-Policy Maximum Entropy Deep Reinforcement Learning with a Stochastic Actor},
  author = {Haarnoja, Tuomas and Zhou, Aurick and Abbeel, Pieter and Levine, Sergey},
  booktitle = {Proceedings of the 35th International Conference on Machine Learning},
  pages = {1861--1870},
  year = {2018},
  editor = {Dy, Jennifer and Krause, Andreas},
  volume = {80},
  series = {Proceedings of Machine Learning Research},
  month = {10--15 Jul},
  publisher = {PMLR},
}

@misc{schulman2017ppo,
    title={Proximal Policy Optimization Algorithms}, 
    author={John Schulman and Filip Wolski and Prafulla Dhariwal and Alec Radford and Oleg Klimov},
    year={2017},
    archivePrefix={arXiv},
    eprint={1707.06347},
    url={https://arxiv.org/abs/1707.06347}, 
}

@article{zhang2025flight,
    title = {Learning vision-based agile flight via differentiable physics},
    author = {Zhang, Yuang and Hu, Yu and Song, Yunlong and Zou, Danping and Lin, Weiyao},
    journal = {Nature Machine Intelligence},
    volume={7},
    number={6},
    pages={954--966},
    year={2025},
}

@inproceedings{liu2026dexndm,
    title={Dex{NDM}: Closing the Reality Gap for Dexterous In-Hand Rotation via Joint-Wise Neural Dynamics Model},
    author={Xueyi Liu and He Wang and Li Yi},
    booktitle={The fourteenth International Conference on Learning Representations},
    year={2026}
}

@inproceedings{zou2025excavator,
  author={Zou, Ziqing and Wang, Cong and Hu, Yue and Liu, Xiao and Xu, Bowen and Xiong, Rong and Fan, Changjie and Chen, Yingfeng and Wang, Yue},
  booktitle={2025 IEEE/RSJ International Conference on Intelligent Robots and Systems (IROS)}, 
  title={High-Precision and High-Efficiency Trajectory Tracking for Excavators Based on Closed-Loop Dynamics}, 
  year={2025},
  volume={},
  number={},
  pages={5617-5624},
  doi={10.1109/IROS60139.2025.11246021}}

@misc{nan2025excavator,
    author = {Nan, Fang and Ma, Hao and Guan, Qinghua and Hughes, Josie and Muehlebach, Michael and Hutter, Marco},
    title = {Efficient Model-Based Reinforcement Learning for Robot Control via Online Learning},
    year = {2025},
    eprint={2510.18518},
    archivePrefix={arXiv},
    url = {https://arxiv.org/abs/2510.18518}
}

@article{nitish2014dropout,
  author  = {Nitish Srivastava and Geoffrey Hinton and Alex Krizhevsky and Ilya Sutskever and Ruslan Salakhutdinov},
  title   = {Dropout: A Simple Way to Prevent Neural Networks from Overfitting},
  journal = {Journal of Machine Learning Research},
  year    = {2014},
  volume  = {15},
  number  = {56},
  pages   = {1929--1958},
}

@misc{ba2016layernormalization,
      title={Layer Normalization}, 
      author={Jimmy Lei Ba and Jamie Ryan Kiros and Geoffrey E. Hinton},
      year={2016},
      eprint={1607.06450},
      archivePrefix={arXiv},
      primaryClass={stat.ML},
      url={https://arxiv.org/abs/1607.06450}, 
}

\end{document}